\def\year{2018}\relax
\documentclass[letterpaper]{article} 
\usepackage{aaai18}  
\usepackage{times}  
\usepackage{helvet}  
\usepackage{courier}  
\usepackage{url}  
\usepackage{amsmath}
\usepackage[noend]{algorithmic}
\usepackage[ruled]{algorithm2e}
\usepackage{amssymb}
\usepackage{CJKutf8}
\usepackage{CJK}
\usepackage{epstopdf}
\usepackage{times}
\usepackage{makecell}
\usepackage{url}
\usepackage{latexsym}
\usepackage{graphicx}
\usepackage{amsmath}
\usepackage{amssymb}
\usepackage{amsmath}
\usepackage{CJK}
\usepackage{graphicx}  
\newlength\myindent
\newcolumntype{C}[1]{>{\centering\let\newline\\\arraybackslash\hspace{0pt}}m{#1}}
\begin{document}
	\begin{CJK*}{UTF8}{gbsn} 
		%
		\title{Neural Response Generation with Dynamic Vocabularies}
		
	\author{	Yu Wu$^{\dag}$, Wei Wu$^\ddag$, Dejian Yang$^\dag$, Can Xu$^\ddag$,Zhoujun Li$^\dag$, Ming Zhou$^\ddag$\\
		$^\dag$State Key Lab of Software Development Environment, Beihang University, Beijing, China\\
		$^\ddag$~~~~Microsoft Research, Beijing, China\\
		\{wuyu,dejianyang,lizj\}@buaa.edu.cn \{wuwei,can.xu,mingzhou\}@microsoft.com 
		}
		\maketitle
		\begin{abstract}
			
			We study response generation for open domain conversation in chatbots. Existing methods assume that words in responses are generated from an identical vocabulary regardless of their inputs, which not only makes them vulnerable to generic patterns and irrelevant noise, but also causes a high cost in decoding. We propose a dynamic vocabulary sequence-to-sequence (DVS2S) model which allows each input to possess their own vocabulary in decoding. In training, vocabulary construction and response generation are jointly learned by maximizing a lower bound of the true objective with a Monte Carlo sampling method. In inference, the model dynamically allocates a small vocabulary for an input with the word prediction model, and conducts decoding only with the small vocabulary.  Because of the dynamic vocabulary mechanism, DVS2S eludes many generic patterns and irrelevant words in generation, and enjoys efficient decoding at the same time. Experimental results on both automatic metrics and human annotations show that DVS2S can significantly outperform state-of-the-art methods in terms of response quality, but only requires 60$\%$ decoding time compared to the most efficient baseline.
			
		\end{abstract}
		\section{Introduction}
		Together with the rapid growth of social conversation data on Internet, there has been a surge of interest on building chatbots for open domain conversation with data driven approaches. Existing methods are either retrieval based \cite{wang2013dataset,yan2016learning,DBLP:conf/acl/WuWXZL17} or generation based \cite{vinyals2015neural,ritter2011data,shang2015genBased}. Recently, generation based approaches are becoming popular in both academia and industry, and a common practice is to learn a response generation model within an encoder-decoder framework (a.k.a., a sequence-to-sequence model) from the large scale conversation data. The mainstream of implementation of the encoder-decoder framework is using neural networks, because they are powerful on capturing complicated semantic and syntactic relations between messages and responses and are end-to-end learnable. On top of the architecture, various models have been proposed to tackle the notorious ``safe reply'' problem \cite{xing2016topic,mou2016sequence,li2015diversity}; to take conversation history into consideration \cite{sordoni2015neural,serban2015building,DBLP:conf/aaai/SerbanSLCPCB17,DBLP:conf/acl/ZhaoZE17}; and to bias responses to some specific persona or emotions \cite{li2016persona,zhou2017emotional}.   
		
		Although existing work has made great progress on generating proper responses, they all assume a static vocabulary in decoding, that is they use the same large set of words to generate responses regardless of inputs. The assumption, however, is a simplification of the real scenario, as proper responses to a specific input (either a message or a conversation context) could only relate to a small specific set of words, and the sets of words could be different from input to input. As a result, the assumption may cause some problems in practice: (1) words that are semantically far from the current conversation also take part in decoding. These words may bias the process of generation and increase the probability of irrelevant responses and generic responses when some of them appear very frequently in the entire data set; (2) the decoding process becomes unnecessarily slow, because one has to estimate a probability distribution for the entire static vocabulary in decoding of each word of a response. More seriously, to suppress the irrelevant responses and the generic responses, state-of-the-art methods have to either complicate their decoders \cite{xing2016topic,mou2016sequence} or append a heavy post-processing procedure after decoding \cite{li2015diversity}, which further deteriorates efficiency. These problems widely exist in the existing methods, but have not drawn enough attention yet.           
		
		In this paper, we aim to achieve high quality response generation and fast decoding at the same time. Our idea is that we dynamically allocate a vocabulary for each input at the decoding stage. The vocabulary is small as it only covers words that are useful in forming relevant and informative responses for the input and filters most irrelevant words out. Because response decoding of each input only focuses on their own relevant words, the process can be conducted efficiently without loss of response quality.  We formulate the idea as a dynamic vocabulary sequence-to-sequence (DVS2S) model. The model defines a dynamic vocabulary in decoding through a multivariate Bernoulli distribution \cite{dai2013multivariate} on the entire vocabulary and factorizes the generation probability as the product of a vocabulary generation probability conditioned on the input and a response generation probability conditioned on both the input and the vocabulary. DVS2S follows the encoder-decoder framework. In encoding, an input is transformed to a sequence of hidden vectors. In decoding, the model first estimates the multivariate Bernoulli distribution using the hidden vectors given by the encoder, and then selects words to form a vocabulary for the decoder according to the  distribution. Responses are generated only using the selected words. Vocabulary construction and response generation are jointly learned from training data, and thus in parameter learning errors in response prediction can be backpropagated to vocabulary formation and used to calibrate word selection. In training, as target vocabularies can only be partially observed from data, we treat them as a latent variable, and optimize a lower bound of the true objective through a Monte Carlo sampling method.   
		
		We conduct an empirical study using the data in \cite{xing2016topic}, and compare DVS2S with state-of-the-art generation methods using extensive automatic evaluation metrics and human judgment. In terms of automatic evaluation, DVS2S achieves $3\%$ gain on BLEU-1 and $5\%$ gain on Embedding Average \cite{liu2016not} over the best performing baseline. On human evaluation, DVS2S significantly outperforms the baseline methods, which is consistent with the automatic evaluation results. Moreover, the model also achieves 6$\%$ gain on the metric of distinct-1 over the best baseline model, indicating that it can generate more informative and diverse responses. Upon the significant improvement on response quality, DVS2S can save 40$\%$ decoding time compared to the most efficient baseline in the same running environment.  
		
		Our contributions in the paper are three-folds: (1) proposal of changing the static vocabulary mechanism to a dynamic vocabulary mechanism in the response generation for chatbots; (2) proposal of a dynamic vocabulary sequence-to-sequence model and derivation of a learning approach that can jointly optimize word selection and response generation;  (3) empirical verification of the effectiveness and efficiency of the proposed model on large scale conversation data.
		
		\section{Related Work}
		Recent years have witnessed remarkable success on open domain response generation for chatbots. In a single-turn scenario, Ritter \cite{ritter2010unsupervised} formulated response generation as a machine translation problem by regarding messages and responses as a source language and a target language respectively. Due to the success on machine translation, sequence-to-sequence (S2S) models \cite{bahdanau2014neural} have been widely used in response generation recently. For instance, Vinyals et al. \cite{vinyals2015neural} and Shang et al. \cite{shang2015genBased} applied S2S with attention on this task.  To address the ``general response'' issue of the standard S2S, Li et al. \cite{li2015diversity} presented a maximum mutual information objective function, and Mou et al. \cite{mou2016sequence} and Xing et al. \cite{xing2016topic} incorporated external knowledge into the S2S model. Shao et al. \cite{shao2017generating} proposed a target attention neural conversation model to generate long and diverse responses.  Reinforcement learning \cite{li2016deep} and adversarial learning \cite{li2017adversarial,xu-EtAl:2017:EMNLP2017} techniques have also been exploited to enhance the existing models. Apart from the effort on improving response quality, researchers also considered varing persona and emotions of generated responses \cite{li2016persona,zhou2017emotional}. In a multi-turn scenario, Sordoni et al. \cite{sordoni2015neural} compressed context information into a vector and injected the vector into response generation. Serban et al. \cite{serban2015building} adopted a hierarchical recurrent structure to model multi-turn conversations. As an extension of the model, latent variables were introduced to model the ``one-to-many'' relation in conversation \cite{DBLP:conf/aaai/SerbanSLCPCB17,DBLP:conf/acl/ZhaoZE17}.  
		
		In this work, we focus on an important but less explored problem: vocabulary selection in decoding. We propose changing the widely used static vocabulary decoder in both single-turn generation and multi-turn generation to a dynamic vocabulary decoder, and derive an approach to jointly learn vocabulary construction and response generation from data.  The proposed method can improve response quality and at the same time speed up decoding process. 
		
		Before us, some work in machine translation has already exploited dynamic vocabularies \cite{l2016vocabulary,DBLP:conf/acl/JeanCMB15,mi2016vocabulary}. These work often treats vocabulary construction and translation as two separate steps. The same practice, however, cannot be easily transplanted to conversation, as there are no clear ``one-to-one'' translation relations in responding. To maintain response quality while improve efficiency in conversation, we propose joint learning of vocabulary construction and response generation in order to let them supervise each other. As far as we know, we are the first who explore the application of dynamic vocabularies in response generation for open domain conversation.   
		
		\begin{figure*}[t]		
			\begin{center}
				\includegraphics[width=12cm,height=6cm]{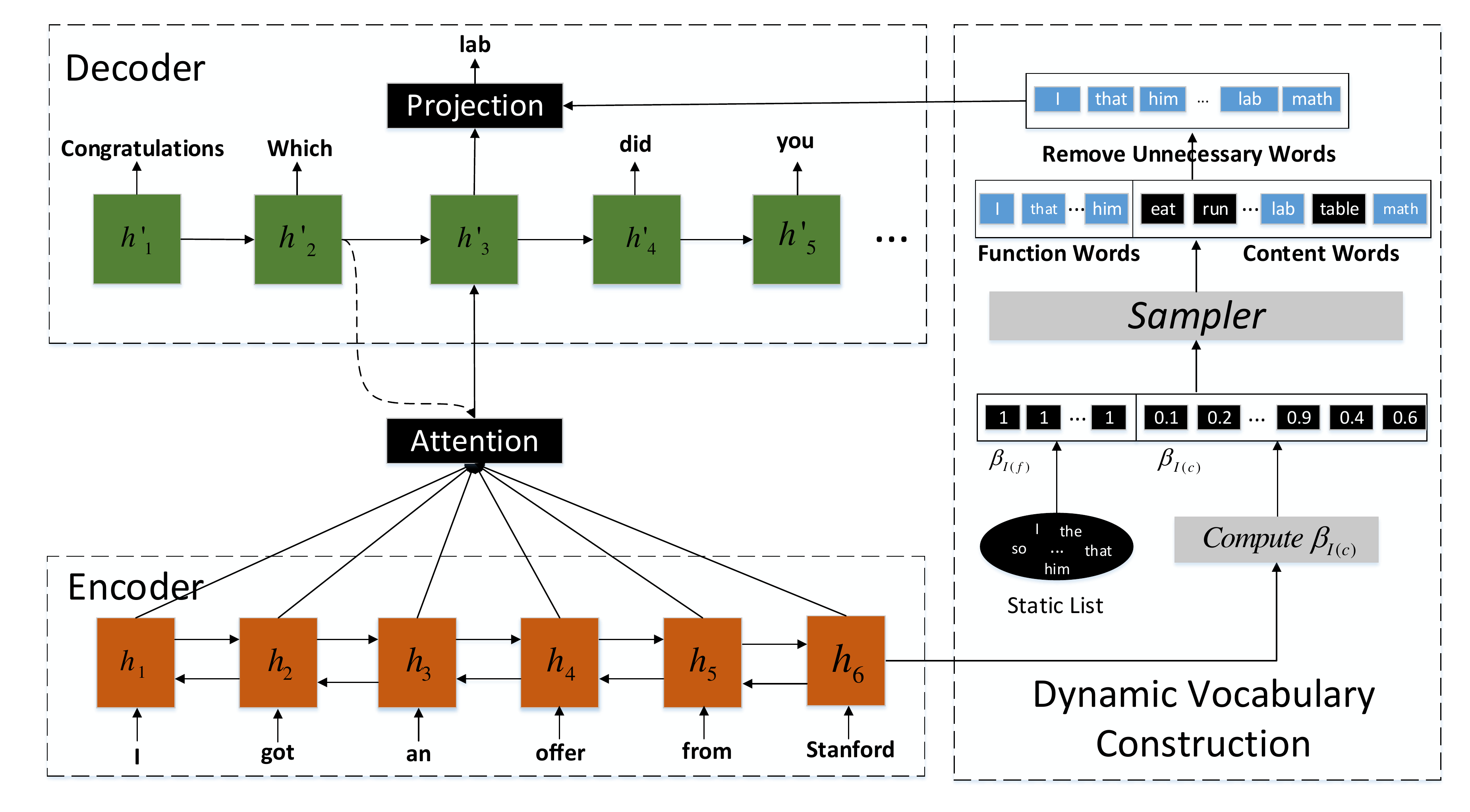}
			\end{center}\vspace{-4mm}
			\caption{Architecture of DVS2S. We translate a Chinese example into English and show the generation process of the word ``lab''. Words in blue squares refers to those chosen by the dynamic vocabulary otherwise they are in black squares.}\label{fig:arch} \vspace{-5mm}
		\end{figure*}
		
		\section{Approach}
		\subsection{Problem Formalization}
		Suppose that we have a data set $\mathcal{D} = \{(X_i,Y_i)\}^N_{i=1} $, where $Y_i$ is a response of an input $X_i$. Here $X_i$ can be either a message or a message with several previous turns as a context. As the first step, we assume $X_i$ a message in this work, and leave the verification of the same technology to context-based response generation as future work. $\forall i$, $X_i$ corresponds to a target vocabulary (i.e., vocabulary in decoding) $T_i = (t_{i,1}, \ldots, t_{i,|V|})$ sampled from a multivariate  Bernoulli distribution $(\beta_{i,1}, \ldots \beta_{i,|V|})$ where $|V|$ is the size of the entire vocabulary $V$ and $t_{i,j} \in \{0,1\}, \forall 1 \leqslant j \leqslant |V|$. $t_{i,j} = 1$ means that the $j$-th word $w_j$ in $V$ is selected for generating responses for $X_i$, otherwise the word will not be used in generation. $\beta_{i,j}=p(t_{i,j} = 1)$ is the probability of the $j$-th word being selected which is parameterized by a function $f(X_i)$. Generation probability of $Y_i$ given $X_i$ is formulated as  $p(Y_i|X_i)= p(Y_i|T_i,X_i)p(T_i|X_i)$.
		
		Our goal is to learn a word selection model $f(X)$ (corresponds to $p(T|X)$) and a response generation model $g(X,T)$ (corresponds to $p(Y|T,X)$) by maximizing log-likelihood $\sum_{i=1}^N \text{log}[p(Y_i|X_i)]$ of $\mathcal{D}$. Thus given a new message $X'$, we can estimate its target vocabulary $T'$ with $f(X')$ and generate a response $Y'$ using $g(X',T')$.  In the following sections, we first introduce our DVS2S model (i.e. $g(X,T)$) by assuming that $T$ is obtained. Then we present how to sample $T$ with the use of  $f(X)$. Finally, we show how to jointly learn $f(X)$ and $g(X,T)$ from $\mathcal{D}$.

		\subsection{Dynamic Vocabulary Sequence-to-Sequence Model}
		Figure \ref{fig:arch}  illustrates the architecture of our dynamic vocabulary sequence-to-sequence (DVS2S) model. DVS2S is built in an encoder-decoder framework \cite{sutskever2014sequence} with an attention mechanism \cite{bahdanau2014neural}. For each input, it equips the decoder with a specific vocabulary that consists of useful words sampled from the entire vocabulary according to a distribution and performs response generation with the vocabulary. Specifically, given a message $X = (x_1,x_2,\ldots, x_t)$ where $x_i$ is the embedding of the $i$-th word, the encoder exploits a bidirectional recurrent neural network with gated recurrent units (biGRU) \cite{chung2014empirical} to transform $X$ into hidden vectors $h = (h_1,h_2,\ldots, h_t)$. A biGRU comprises a forward GRU that reads a sentence in its order and a backward GRU that reads the sentence in its reverse order. The forward GRU encodes the sentence into hidden vectors $(\overrightarrow{h}_1, \ldots, \overrightarrow{h}_t)$ by
		\begin{eqnarray} \small
		&& z_i = \sigma(W_z x_i + U_z  \overrightarrow{h}_{i-1}),  \\ \nonumber
		&& r_i = \sigma(W_r  x_i + U_r  \overrightarrow{h}_{i-1}), \\ \nonumber
		&&\widetilde{h}_i = tanh(W_h x_i + U_h (r_i \odot \overrightarrow{h}_{i-1})),\\ \nonumber
		&&\overrightarrow{h}_i = z_i \odot \widetilde{h}_i + (1-z_i) \odot \overrightarrow{h}_{i-1},
		\end{eqnarray}
		where $z_i$ and $r_i$ are an update gate and a reset gate respectively, $\overrightarrow{h}_{0}=0$, and $W_z$, $W_h$, $W_r$, $U_z$, $U_r$,$ U_h $ are parameters. The backward hidden state $\overleftarrow{h}_i$ is obtained similarly.
		Then $\forall i \in [1,t]$, $h_{i}$ is the concatenation of $\overrightarrow{h}_i$ and  $\overleftarrow{h}_i$. 
		
		The decoder takes $h = (h_1,h_2,\ldots, h_t)$ as an input and generates a response by a language model with an attention mechanism. When generating the $i$-th word $y_i$, the decoder estimates a word distribution $\hat{y_i}$ by
		\begin{equation}\small
		\hat{y_i} =l(y_{i-1},c_i,h'_i, T),
		\end{equation} 
		where $c_i$ is a context vector formed by the attention mechanism, $h'_i$ is the $i$-th hidden state of the decoder, and $y_{i-1}$ is the $(i-1)$-th word of the response. Specifically, the decoder also exploits a GRU to encode $y_{i-1}$ into $h'_i$ whose initial state is the last hidden vector of the encoder.  $c_i$ is a linear combination of $\{h_1, \ldots, h_t\}$ which is formulated as
		\begin{equation}\small
		c_i = \sum_{j=1}^t \alpha_{i,j} h_j,
		\end{equation}
		where $\alpha_{i,j}$ is given by 
		\begin{eqnarray}\small
		&& \alpha_{i,j} = \frac{exp(e_{i,j})}{\sum_{k=1}^t exp(e_{i,k})}, \\
		&& e_{i,j} = v^\top tanh(W_{\alpha}[h_j;h'_i]).
		\end{eqnarray}
		$W_{\alpha}$ and $v$ are parameters, and $[\cdot ; \cdot]$ means concatenation of the two arguments. $l(y_{i-1},c_i,h'_i, T)$ is a $|T|$-dimensional probability distribution where $|T|=\sum_{k=1}^{|V|} t_k$. $\forall t_k \in T$, if $t_k=1$, then the corresponding element in 
		$l(y_{i-1},c_i,h'_i, T)$ is defined by
		\begin{equation} \small  \label{softmax}
		p(y_i=w_k)=\frac{exp(s(w_k))}{\sum_{t_j \in T, t_j=1} exp(s(w_j))},
		\end{equation}	
		where $s(w_k)$ is given by
		\begin{equation}\small \label{dynamic_equation}
		s(w_k) = W_{w_k} [y_{i-1} ; h'_{i-1} ; c_i] + b_{w_k}, \forall t_k \in T.
		\end{equation}
		$W_{w_k}$ and $b_{w_k}$ are two parameters. Equation (\ref{softmax}) and (\ref{dynamic_equation}) are called projection operation. 
		
		
		Time complexity of decoding of DVS2S is $\mathcal{O}(  len_r\cdot m \cdot p + len_r \cdot len_m \cdot m^2+ len_r \cdot(m+p)\cdot |T| + m \cdot |V|)$ (GRU+attention+projection+vocabulary construction), while time complexity of decoding of the existing methods is at least  $\mathcal{O}(  len_r\cdot m \cdot p + len_r \cdot len_m \cdot m^2+ len_r \cdot (m+p)\cdot |V|)$ (GRU+attention+projection), where $len_r$ is the length of the generated response, $len_m$ is the length of the message, $m$ is the hidden state size of the decoder, and $p$ is the embedding size of target words. In practice, $|V|$ is much larger than other parameters, so the cost of decoding in existing methods is dominated by $len_r \cdot (m+p)\cdot |V|$ (i.e., time complexity of projection). DVS2S reduces it to  $len_r \cdot  (m+p)\cdot |T|$ in Equation (\ref{softmax}) and (\ref{dynamic_equation}). Since $len_r$ is usually much larger than $1$, $len_r \cdot(m+p)\cdot |T| + m \cdot |V|$ is much smaller than $len_r \cdot(m+p)\cdot |V|$. Therefore, DVS2S could enjoy a faster decoding process than the existing methods (the conclusion is also verified in experiments). 

		\subsection{Dynamic Vocabulary Construction}\label{dvcons}
		In this section, we elaborate dynamic vocabulary construction for $X$. We define $\overline{T}=\{w_k \in V | t_k \in T, t_k=1\}$ and $I(w)$ the index of word $w$ in $V$. $\overline{T}$ is equivalent to $T$. Remember that $T$ is a variable sampled from a multivariate Bernoulli distribution which is a joint distribution of $|V|$ independent Bernoulli distributions. Each Bernoulli distribution depicts the probability of a word $w$ from $V$ being selected to $\overline{T}$ and is parameterized by $\beta_{I(w)}$. We make such an assumption because there does not exist a clear ``one-to-one'' relationship between words in a message and words in its proper responses and we have to treat $T$ as a latent variable in training as useful words for forming a proper response to a message can only be partially observed in training data. 
		
		$\overline{T} =\overline{T}_c \cup \overline{T}_f$ where $\overline{T}_c$ refers to content words and $\overline{T}_f$ refers to function words. Function words guarantee grammatical correctness and fluency of responses. Therefore, there should not be a large variance on $\overline{T}_f$ over difference messages. We collect words appearing more than $10$ times in the training data, excluding nouns, verbs, adjectives and adverbs from them, and use the remaining ones to form a function word set $\overline{V}_f$ of $V$. $\forall w \in \overline{V}_f$, we define $\beta_{I(w)}=1$. Thus, $\overline{T}_f=\overline{V}_f$ regardless of inputs. In other words, all function words are always sampled in the construction of $\overline{T}$.  
		
		Content words, on the other hand, express semantics of responses, and thus should be highly related to the input message. Let $\overline{V}_c = \overline{V} \setminus \overline{V}_f$ be the full content word set, then $\forall c \in \overline{V}_c$, we parameterize $\beta_{I(c)}$ as 
		\begin{equation}\label{wordpredict}\small
		\beta_{I(c)} = \sigma(W_c^{\top} h_t + b_c),
		\end{equation}
		where $\sigma$ is a sigmoid function, $h_t$ is the last hidden state of the encoder, and $W_c$ and $b_c$ are parameters. In the construction of $\overline{T}$, $\overline{T}_c$ is sampled from $\overline{V}_c$ based on $\{\beta_{I(c)} | c \in \overline{V}_c\}$.
		
		How to allocate a proper $\overline{T}$ to $X$ is key to the success of DVS2S. $\overline{T}$ should cover enough words that are necessary to generate relevant, informative, and fluent responses for $X$, but cannot be too large for the sake of cost control in decoding. To make sure that we can sample such a $\overline{T}$ with high probability, we consider jointly learning vocabulary construction and response generation from training data, as will be seen in the next section.      
		
		\subsection{Model Training}
		With a latent variable  $T$, the objective of learning can be written as 
		\begin{equation} \label{trueobj}
		\sum_{i=1}^N	\log(p(Y_i|X_i)) =  \sum_{i=1}^N \log(\sum_{T_i} p(Y_i|T_i,X_i)p(T_i|X_i)).
		\end{equation}
		Equation (\ref{trueobj}) is difficult to optimize as logarithm is outside the summation. Hence, we instead maximize a variational lower bound of $ \sum_{i=1}^N	\log[p(Y_i|X_i)]$ which is given by  
		\begin{eqnarray} \small \label{obj}
		L && =  \sum_{i=1}^N \sum_{T_i} p(T_i|X_i) \log p(Y_i|T_i,X_i) \\
		&& =  \sum_{i=1}^N  \sum_{T_i} \big[ \prod_{j=1}^{|V|} p(t_{i,j}|X_i) \sum_{l=1}^m \log p(y_{i,l}|y_{i,<l}, T_i,X_i)\big] \nonumber \\ 
		&&	\leq \sum_{i=1}^N   \log(\sum_{T_i} p(Y_i|T_i,X_i)p(T_i|X_i)) \nonumber \\ 
		&&	=  \sum_{i=1}^N  \log[ p(Y_i|X_i)]\nonumber 
		\end{eqnarray}
		Let $\Theta$ represent the parameters of $L$ and $\frac{\partial L_i(\Theta)}{\partial \Theta}$ be the gradient of $L$ on an example $X_i \in \mathcal{D}$, then $\frac{\partial L_i(\Theta)}{\partial \Theta}$ can be written as 
		\begin{equation} \label{grad} \small
		\sum_{T_i} p(T_i|X_i)\big[ \frac{\partial  \text{log} p(Y_i|T_i,X_i)}{\partial \Theta}  + 
		\text{log} (Y_i|T_i,X_i) \frac{\partial \text{log}  p(T_i|X_i)}{\partial \Theta}  \big]
		\end{equation}
		Enumerating all $2^{|V|}$ samples of $T_i$ in Equation (\ref{grad}) is intractable. Therefore, we employ the Monte Carlo sampling technique to approximate $\frac{\partial L_i(\Theta)}{\partial \Theta}$. Suppose that we have $S$ samples, then the approximation of the gradient can be written as 
		\begin{equation} \small
		\frac{1}{S} \sum_{s=1}^S \big[ \frac{\partial  \text{log} p(Y_i|\widetilde{T}_{i,s},X_i)}{\partial \Theta}  + 
		\text{log} (Y_i|\widetilde{T}_{i,s},X_i) \frac{\partial \text{log}  p(\widetilde{T}_{i,s}|X_i)}{\partial \Theta}  \big],
		\end{equation} 
		where $\widetilde{T}_{i,s} \sim \text{a multivariate Bernoulli distribution} (\{\beta_i\}^{|V|}) $. To reduce variance, we normalize the gradient with the length of the response and introduce a moving average baseline $b_k$ to the gradient \cite{weaver2001optimal}:	 		
		\begin{equation}\label{gradient} \small
		\begin{aligned}
		&\frac{\partial L_i(\Theta)}{\partial \Theta}  \approx \frac{1}{S} \sum_{s=1}^S \big[ \frac{\partial  \text{log} p(Y_i|\widetilde{T}_{i,s},X_i)}{\partial \Theta} \\& + 
		((\frac{1}{m} \sum_{j=1}^m \text{log} p(y_{i,j}|y_{i,< j}, |\widetilde{T}_{i,s},X_i)- b_{k}) \frac{\partial \text{log}  p(\widetilde{T}_{i,s}|X_i)}{\partial \Theta}  \big],
		\end{aligned}
		\end{equation} 		
		where $b_k$ is the baseline after $k$-th mini-batch, and $b_k$ is updated using the following equation:
		\begin{equation}\label{b_k} \small
		b_{k+1} = 0.9\times b_{k} +\frac{0.1}{mS} \sum_{Y_i \in \text{batch } k} \sum_{s=1}^S \sum_{j=1}^m \text{log} [p(y_{i,j}|y_{i,<j}, \widetilde{T}_{is},X_i)].
		\end{equation}
		
		We summarize our training algorithm in Algorithm \ref{alg} where we initialize $\Theta$ by pre-training an S2S model and a word prediction model to facilitate convergence and use a mini-batch training strategy to update it and the baseline $\{b_k\}$. We employ AdaDelta algorithm \cite{zeiler2012adadelta} to train our model with a batch size $64$. We set the initial learning rate as $1.0$ and reduce it by half if  perplexity on validation begins to increase. We will stop training if the perplexity on validation keeps increasing in two successive epochs.
		\begin{algorithm}[th]\small
			\caption{Optimization Algorithm }
			\SetKwData{Index}{Index}
			\SetKwInput{kwInit}{Init}
			\SetKwInput{kwOutput}{Output}
			\SetKwInput{kwInput}{Input}
			\label{alg}
			\small{
				
				\kwInput{ $\mathcal{D}$, $V$, initial learning rate $lr$, MaxEpoch\\}
				\kwInit{ $\Theta$}
				\Indp
				
				Pretrain a Seq2Seq model with $\mathcal{D}$. \\
				
				Fix the encoder, and pre-train $\{W_c, b_c\}$ in Equation (\ref{wordpredict}) by maximizing $\sum_{i=1}^{N}\sum_{j=1}^{|V|} \log[p(t_{i,j}|X_i)]$\\
				\Indm
				\While{e $<$ \text{MaxEpoch} \textbf{and}\xspace perplexity does not increase in 2 successive epchos}
				{
					\ForEach{mini-batch k }
					{
						Compute the sampling probability $\{\beta_i\}^{|V|}$ with Equation (\ref{wordpredict})\\
						\For{ s $<$ S}
						{
							Sample a $\widetilde{T}_s \sim \text{multivariate Bernoulli} (\{\beta_i\}^{|V|})$\\
							Compute loss according to Equation (\ref{obj}) \\
							Compute gradient according to Equation (\ref{gradient})
						}
						Update $b_{k}$ according to Equation (\ref{b_k}) \\
						Update parameter $\Theta$ with AdaDelta algorithm	 					
					}
					\If{perplexity increases}{$lr = lr /2 $}
				}

				\kwOutput{  $\Theta$}
			}
			
		\end{algorithm}
		One advantage of joint learning is that errors in response prediction in training can be backpropagated to vocabulary construction and signals from response can help calibrate word selection. Therefore, the learning approach can mitigate discrepancy between training and inference in practice.  It is easy to extend DVS2S to handle multi-turn response generation by replacing its encoder with one that can model contexts (e.g., the one in \cite{serban2015building}), and model learning can also be enhanced using techniques like adversarial learning \cite{li2017adversarial} and reinforcement learning \cite{li2016deep} by re-defining the objective function in (\ref{obj}).   
		\section{Experiment}
		We compare DVS2S with state-of-the-art response generation models in terms of both efficacy and efficiency.  
		\subsection{Experiment Setup}
		We use the data in \cite{xing2016topic} which consists of message-response pairs crawled from Baidu Tieba\footnote{\url{https://tieba.baidu.com/}}. Messages and responses are tokenized by Standford Chinese word segmenter.
		There are $5$ million pairs in the training set, $10,000$ pairs in the validation set, and $1,000$ pairs in the test set. Messages in the test data are used to generate responses, and responses in the test data are treated as ground truth to calculate automatic evaluation metrics. Both the message vocabulary and the response vocabulary contain $30,000$ words that cover $98.8\%$ and $98.3\%$ of words appearing in the messages and in the responses respectively in the training data. In this work, we take the response vocabulary in the data as the entire vocabulary for decoding (i.e., $V$). 
		
		We implement our model using Theano \cite{2016arXiv160502688short}. In our model, we set the word embedding size as $620$ and the hidden vector size as $1024$ in both encoding and decoding. In the Monte Carlo sampling, we set the number of samples $S$ as $5$. We follow the method described in the dynamic vocabulary construction section to construct target vocabularies. There are $701$ function words. In test, we rank content words according to $\{\beta_i\}$ and select top $1,000$ words to form a target vocabulary for a message with the function words. This is equivalent to sampling many times and selecting top $1,000$ words according to their frequency in the union of all samples. The strategy does not change the time complexity of decoding and could reduce variance of the model in inference. We set the beam size as $20$ and use the top one response from beam search in evaluation. Code will be released later.
		\begin{table*}[t]
			\small
			\centering
			\caption{Automatic evaluation results. Numbers in bold mean that improvement from the model on that metric is statistically significant over the baseline methods (t-test, p-value $<0.01$). 
			}		\label{exp} 	
			
			\begin{tabular}{l|c|c|c|c|c|c|c|c}
				\hline
				& BLEU-1 & BLEU-2 & BLEU-3&  Average&  Extrema &  Greedy &  Distinct-1 &  Distinct-2\\ \hline
				
				S2SA & 4.96 & 1.96 & 0.81 & 25.32&11.70& 24.73& 0.091 & 0.207\\ 
				S2SA-MMI  & 5.82 & 1.47 & 0.70 &27.16& 14.96& 25.89 & 0.151 & 0.378\\ 
				TAS2S &  6.26 & 2.11 & 0.98 &27.92&15.86&26.29&0.161 & 0.401\\ 	 		 		
				CVAE &  6.33 & 1.86 & 0.55 &28.92&15.01&26.13 & 0.143 & 0.346\\ \hline
				S-DVS2S & \textbf{8.01}&2.94&0.93&\textbf{32.41}&\textbf{20.15}&\textbf{29.89} & \textbf{0.221} &\textbf{ 0.601}\\ 
				DVS2S &\textbf{9.89}&3.51&1.33&\textbf{34.05}&\textbf{22.72}&\textbf{31.61}& \textbf{0.233} & \textbf{0.632}\\ \hline
				
				\hline
			\end{tabular}	
		\end{table*}
		\subsection{Evaluation Metrics}
		we evaluate the performance of different models with the following metrics:  
		
		\textbf{Word overlap based metrics\footnote{As our model makes prediction on a small vocabulary, perplexity is not a proper metric for evaluation.}}: following previous work \cite{li2015diversity,tian2017make}, we employ BLEU-1, BLEU-2, and BLEU-3 as evaluation metrics.  
		
		\textbf{Embedding based metrics}: following  \cite{serban2015building,DBLP:conf/acl/ZhaoZE17}, we employ Embedding Average (Average), Embedding Extrema (Extrema), and Embedding Greedy (Greedy) as evaluation metrics. These metrics are based on word embeddings, and they can measure  relevance of a response regarding to a message when there is little word overlap between them. According to Liu et al. \cite{liu2016not}, these metrics have higher correlation with human judgment than BLEUs. We obtain word embeddings by running a public word2vec tool\footnote{\url{https://code.google.com/archive/p/word2vec/}} on the $5$ million training data. The embedding size is set as $200$.   
		
		\textbf{Distinct-1 \& distinct-2}: following \cite{li2015diversity,xing2016topic}, we calculate the ratios of distinct unigrams and bigrams in generated responses, and use the metrics to measure how diverse and informative the responses are. 
		
		\textbf{3-scale human annotation}: in addition to the automatic metrics, we recruit three human annotators with rich Tieba experience to judge the quality of the generated responses. Responses from different models are pooled and randomly shuffled for each annotator. Each response is rated by the three annotators under the following criteria: \textbf{+2}: the response is not only relevant and natural, but also informative and interesting; \textbf{+1}: the response can be used as a reply to the message, but might not be informative enough (e.g., “Yes, I see” , “Me too”, and “I don’t know”); \textbf{0}: The response makes no sense, irrelevant, or grammatically broken. 
		
		\subsection{Comparison Methods}
		
		\textbf{S2SA}: the standard S2S model with an attention mechanism \cite{vinyals2015neural}. We use the implementation with Blocks \url{https://github.com/mila-udem/blocks}.
		
		\textbf{S2SA-MMI}: the model proposed by Li et al. \cite{li2015diversity}. We implement this baseline by the code published by the authors at \url{https://github.com/jiweil/Neural-Dialogue-Generation}.
		
		\textbf{TA-S2S}: the topic-aware sequence-to-sequence model proposed in \cite{xing2016topic}. We implement this baseline by the code published by the authors at \url{https://github.com/LynetteXing1991/TAJA-Seq2Seq}.

		\textbf{CVAE}: recent work for response generation with a conditional variational auto-encoder \cite{DBLP:conf/acl/ZhaoZE17}. We use the published code at \url{https://github.com/snakeztc/NeuralDialog-CVAE}
		
		In all the baseline models, we set the parameters as suggested by the existing papers. In addition to these methods, we also compare DVS2S with a simple version of the model. Following \cite{weng2017neural}, we separately learn a generation model and a word prediction model for target vocabulary construction. The procedure is the same as the parameter initialization step in Algorithm \ref{alg}.  The model shares the same embedding size, hidden vector size, target vocabulary size, and the inference process with DVS2S, but differs from DVS2S in that signals from response prediction in training cannot be backpropagated to word prediction for vocabulary construction. We denote the model as \textbf{S-DVS2S}.

		\subsection{Evaluation Results}
		Table \ref{exp} shows the evaluation results on automatic metrics. DVS2S and S-DVS2S significantly outperform the baseline methods on most metrics, demonstrating the effectiveness of the dynamic vocabulary mechanism on response generation for open domain dialogues. Moreover, DVS2S also significantly improves upon S-DVS2S on metrics except BLEU-2 and BLEU-3. The results verify the advantage of joint learning of vocabulary and generation. DVS2S is significantly better than all baseline methods on distinct-1 and distinct-2, indicating that the model can generate more diverse and informative responses. This is because with the dynamic vocabulary mechanism, the model can circumvent the influence from generic patterns when frequent but irrelevant nouns, verbs, adjectives, and adverbs are excluded from decoding, and pay more attention to useful content words in decoding.

		Table \ref{exp:human} reports human evaluation results. DVS2S generates much more informative and interesting responses ($2$ responses) and much less invalid responses ($0$ responses) than the baseline methods. The results are consistent with the automatic evaluation results.  S-DVS2S is much worse than DVS2S on $0$ responses. This is because the gap between training and test in S-DVS2S leads to more grammatical broken and irrelevant responses. Fleiss' Kappa \cite{fleiss1973equivalence} on all models are around $0.4$, indicating relatively high agreement among labelers.  We also conduct a t-test between DVS2S  and the baseline models and results show that the improvement from our model is statistically significant (p-value $<0.01$). 
		
		\begin{table}[h]
			\centering\vspace{-5mm}
			\caption{Human evaluation results. The ratios are calculated by combining the annotations from the three judges together.
			}		\label{exp:human} 	
			
			\begin{tabular}{l|c|c|c|c}
				\hline
				& 0 & 1 & 2 & Kappa \\ \hline
				
				S2SA & 0.321 & 0.564 & 0.115 & 0.43\\ 
				S2SA-MMI  & 0.302 & 0.555 & 0.143 & 0.42\\ 
				TAS2S &  0.249 & 0.571 & 0.180 &0.40 \\ 	 		 		
				CVAE & 0.252 & 0.563 & 0.185 & 0.41 \\ \hline
				S-DVS2S &0.232&0.484&0.284 & 0.38\\ 
				DVS2S &0.094&0.581&0.325 & 0.45\\ \hline
				
				\hline
			\end{tabular}	
		\end{table}

\begin{figure}[h]		
			\begin{center} 
				\includegraphics[width=7cm,height=3cm]{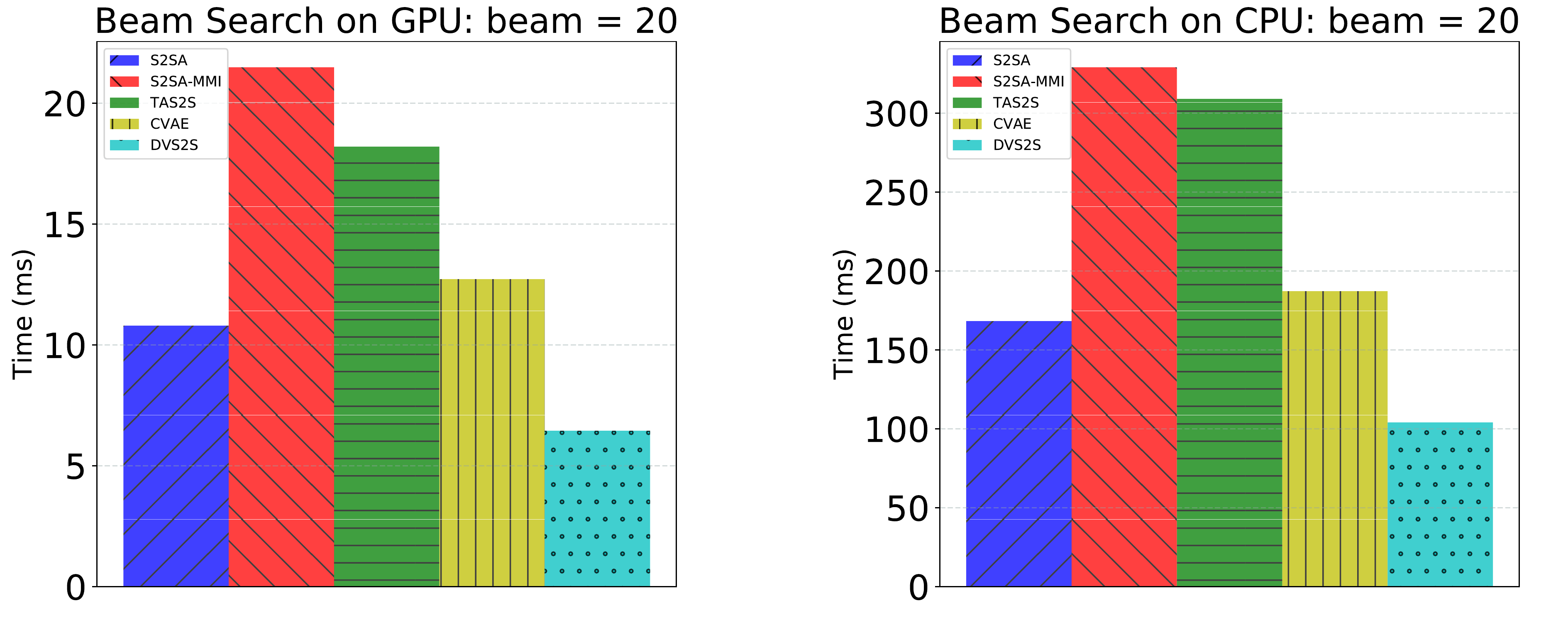}
			\end{center}
			\vspace{-3mm}
			\caption{Efficiency comparison
			}\label{fig:recall}\vspace{-3mm}
		\end{figure}

		In addition to response quality, we also compare DVS2S with baselines on efficiency of decoding. We calculate the average time per word in generating responses for the test messages with a beam size $20$. To make sure that the efficiency comparison is conducted under the setting with which all baselines achieve their best performance on response quality, we use the published codes and the parameters suggested by their papers. S2SA, TAS2S and DVS2S are all implemented on top of Theano, so comparison among them is fair. S2SA-MMI is implemented with Torch, and CVAE is implemented with Tensorflow. Their efficiency might be influenced by the implementation libraries, but they are theoretically not faster than S2SA. We also show their efficiency for reference. The efficiency comparison is conducted on both a GPU environment with a single Tesla K80  and a CPU environment with 6 Intel Xeon CPUs E5-2690 @ 2.6GHz. Figure \ref{fig:recall} gives the comparison results.  We can see that because of the small target vocabularies, DVS2S can save  $40\%$ decoding time on both environments compared to S2SA. TAS2S is better than S2SA on response quality, but it sacrifices efficiency. From the comparisons on both efficiency and efficacy, we can conclude that  DVS2S can achieve  high quality response generation and fast decoding at the same time. 	
		\begin{table*}[t]
			\small
			\centering
			\caption{Case Study. We translate Chinese examples to English, and list top three content words according to $\beta_{I(c)}$.  
			}		\label{case} 	
			
			\begin{tabular}{C{4cm}|C{5cm}|C{3.5cm}|C{3cm}}
				\hline
				Message & DVS2S & TAS2S & S2SA\\	\hline
				
				\begin{tabular}{l}
					我买了辆新奥迪。\\
					I bought a new Audi. \\
				\end{tabular}
				
				&\begin{tabular}{c}
					奥迪的性价比很高。\\
					Audi is cost-effective. \\
					\textbf{keyword}: 奥迪(Audi), \\奔驰(Benz), 价格(price) ...\\
				\end{tabular}  
				& \begin{tabular}{c}
					哈哈，你赢了\\
					haha, you win. \\
				\end{tabular}  & 
				\begin{tabular}{c}
					在哪儿\\
					Where is it? \\
				\end{tabular} 
				\\	\hline
				
				\begin{tabular}{l}
					奢侈品销售收入还不错。\\
					Luxury salesmen have a\\ good payment. \\
				\end{tabular}
				
				&\begin{tabular}{c}
					做销售的都收入不高。\\
					Salesmen do not have high salary. \\
					\textbf{keyword}: 销售(Salesman),\\ 薪水(Salary),金钱(Money) ...\\
				\end{tabular}  
				& \begin{tabular}{c}
					有什么推荐的吗？\\
					Do you have something \\to recommand? \\
				\end{tabular}  & 
				\begin{tabular}{c}
					有兴趣的话可以\\来看看\\
					Come here if\\ you want.  \\
				\end{tabular} 
				\\	\hline
				\begin{tabular}{l}
					为什么这个药店被敲诈了？\\
					Why the pharmacy was\\ extorted? \\
				\end{tabular}
				
				&\begin{tabular}{c}
					药店有卖假货的。\\
					The pharmacy sold fake products. \\
					\textbf{keyword}:药店(pharmacy),\\药(medicine),发烧(fever) ...\\
				\end{tabular}  
				& \begin{tabular}{c}
					我也有这样的问题\\
					I have the same question. \\
				\end{tabular}  & 
				\begin{tabular}{c}
					这是个问题\\
					That is a problem. \\
				\end{tabular} 
				\\	
				\hline
			\end{tabular}	
		\end{table*}
		
		\subsection{Discussions}
		In this section, we give more analysis on DVS2S to help others understand the model. 
		
		\textbf{Dynamic vocabulary coverage.} The first problem we investigate is how many words from the ground truth responses (i.e., responses from human) in the test data are covered by the vocabularies allocated by our algorithm in inference. We measure the coverage by this metric:
		\begin{equation} \small
		Recall = \frac{1}{N_t} \sum_{i=1}^{N_t} \frac{|\{ w | w \in \overline{T}_i \wedge w \in Y_i \}|}{|\{w | w \in Y_i \}|},
		\end{equation} 
		where $N_t$ is the number of instances in the test set (i.e., $1000$), $w$ represents a word, $Y_i$ is the ground truth response in the $i$-th instance, $\overline{T}_i$ is the target vocabulary predicted by DVS2S, and $|\cdot|$ means the number of elements in a set. 
		
		\begin{table}[h]\small
			\vspace{-3mm}
			\centering
			\caption{Ground truth word coverage.
			}		\label{recall} 	
			
			\begin{tabular}{c|c|c|c|c|c|c}
				\hline
				$N$	& 0 & 100 & 1k&  3k&  5k &  10k \\ \hline	 				
				Recall & 63.02 & 72.25 & 79.54 & 85.21&88.39& 92.60\\ 	 						
				\hline
			\end{tabular}	
		\end{table}
		
		Table \ref{recall} reports the metrics varying with respect to the number of content words in the target vocabularies (note that all vocabularies share the same function words). We can see that when selecting top $1000$ content words, the target vocabularies on average can cover about $80\%$ words appearing in the ground truth responses, which is a good balance between efficiency and efficacy. The numbers in the table indicate that useful words can be accurately predicted by the word selection model in DVS2S, and the learning approach generalizes well on the test data. The results are also consistent with the good performance of DVS2S on BLEUs.

		\textbf{Performance across different dynamic vocabulary sizes.} Next, we examine how the performance of DVS2S changes with respect to the size of the target vocabularies. We vary the number of content words selected from the entire vocabulary according to $\{\beta_i\}$ in a range of $\{0,100,1000,3000,5000,10000\}$, and then check how the embedding based metrics change on the test data. Table \ref{performance} shows the results. The results are consistent with our intuition: we may lose important words for response generation when the number of selected words is too small (e.g., less than $100$), but we cannot let the target vocabulary become too large either (e.g., larger than $1000$) because that may involve many irrelevant words into generation. $1000$ is the best choice as the performance of the model reaches its peak.

		%
		
		\begin{table}[h]
			\small
			\centering \vspace{-3mm}
			\caption{Performance in terms of content word number.
			}		\label{performance}
			
			\begin{tabular}{c|c|c|c|c|c|c}
				\hline
				$N$	& 0 & 100 & 1k&  3k&  5k &  10k \\ \hline	 				
				
				Average & 23.43 & 33.87& 34.05& 29.56&27.63& 26.30\\ 	 	
				Greedy  & 23.20& 30.76& 31.61& 28.91&27.32& 25.29\\ 	
				Extrema & 10.20 & 22.51& 22.72& 16.89&14.32& 12.34\\ 	 							
				\hline
			\end{tabular}	
		\end{table}
		
		\textbf{Case study.} Finally, we qualitatively analyze DVS2S with some examples from the test data given in Table \ref{case}. In each example, we also list the top three content words according to the estimated multivariate Bernoulli distribution under the response of our model. Because our model can focus on high quality content words given by the word prediction module in decoding, it can avoid safe responses (e.g., Case 3) and promote responses that are more informative (e.g., Case 1) and more relevant (e.g., Case 2) to top position in beam search of decoding.

		\section{Conclusion and Future Work}
		We consider dynamically allocating a vocabulary to an input in the decoding stage for response generation in open domain conversation. To this end, we propose a dynamic vocabulary sequence-to-sequence model, and derive a learning approach that can jointly optimize vocabulary construction and response generation through a Monte Carlo sampling method. Experimental results on large scale conversation data show that DVS2S can significantly outperform state-of-the-art methods in terms of response quality and at the same time accelerate the decoding process. In the future, we will investigate how to apply the dynamic vocabulary technique to multi-turn response generation, and examine if techniques like reinforcement learning and adversarial learning can further enhance the model.

		\newpage
		\bibliographystyle{aaai}
		\bibliography{acl2017}

\begin{thebibliography}{}

\bibitem[\protect\citeauthoryear{Bahdanau, Cho, and
  Bengio}{2014}]{bahdanau2014neural}
Bahdanau, D.; Cho, K.; and Bengio, Y.
\newblock 2014.
\newblock Neural machine translation by jointly learning to align and
  translate.
\newblock {\em arXiv preprint arXiv:1409.0473}.

\bibitem[\protect\citeauthoryear{Chung \bgroup et al\mbox.\egroup
  }{2014}]{chung2014empirical}
Chung, J.; Gulcehre, C.; Cho, K.; and Bengio, Y.
\newblock 2014.
\newblock Empirical evaluation of gated recurrent neural networks on sequence
  modeling.
\newblock {\em arXiv preprint arXiv:1412.3555}.

\bibitem[\protect\citeauthoryear{Dai \bgroup et al\mbox.\egroup
  }{2013}]{dai2013multivariate}
Dai, B.; Ding, S.; Wahba, G.; et~al.
\newblock 2013.
\newblock Multivariate bernoulli distribution.
\newblock {\em Bernoulli} 19(4):1465--1483.

\bibitem[\protect\citeauthoryear{Fleiss and
  Cohen}{1973}]{fleiss1973equivalence}
Fleiss, J.~L., and Cohen, J.
\newblock 1973.
\newblock The equivalence of weighted kappa and the intraclass correlation
  coefficient as measures of reliability.
\newblock {\em Educational and psychological measurement} 33(3):613--619.

\bibitem[\protect\citeauthoryear{Jean \bgroup et al\mbox.\egroup
  }{2015}]{DBLP:conf/acl/JeanCMB15}
Jean, S.; Cho, K.; Memisevic, R.; and Bengio, Y.
\newblock 2015.
\newblock On using very large target vocabulary for neural machine translation.
\newblock In {\em ACL 2015, July 26-31, 2015, Beijing, China, Volume 1: Long
  Papers},  1--10.

\bibitem[\protect\citeauthoryear{L'Hostis, Grangier, and
  Auli}{2016}]{l2016vocabulary}
L'Hostis, G.; Grangier, D.; and Auli, M.
\newblock 2016.
\newblock Vocabulary selection strategies for neural machine translation.
\newblock {\em arXiv preprint arXiv:1610.00072}.

\bibitem[\protect\citeauthoryear{Li \bgroup et al\mbox.\egroup
  }{2015}]{li2015diversity}
Li, J.; Galley, M.; Brockett, C.; Gao, J.; and Dolan, B.
\newblock 2015.
\newblock A diversity-promoting objective function for neural conversation
  models.
\newblock {\em arXiv preprint arXiv:1510.03055}.

\bibitem[\protect\citeauthoryear{Li \bgroup et al\mbox.\egroup
  }{2016a}]{li2016persona}
Li, J.; Galley, M.; Brockett, C.; Gao, J.; and Dolan, B.
\newblock 2016a.
\newblock A persona-based neural conversation model.
\newblock {\em arXiv preprint arXiv:1603.06155}.

\bibitem[\protect\citeauthoryear{Li \bgroup et al\mbox.\egroup
  }{2016b}]{li2016deep}
Li, J.; Monroe, W.; Ritter, A.; Galley, M.; Gao, J.; and Jurafsky, D.
\newblock 2016b.
\newblock Deep reinforcement learning for dialogue generation.
\newblock {\em arXiv preprint arXiv:1606.01541}.

\bibitem[\protect\citeauthoryear{Li \bgroup et al\mbox.\egroup
  }{2017}]{li2017adversarial}
Li, J.; Monroe, W.; Shi, T.; Ritter, A.; and Jurafsky, D.
\newblock 2017.
\newblock Adversarial learning for neural dialogue generation.
\newblock {\em arXiv preprint arXiv:1701.06547}.

\bibitem[\protect\citeauthoryear{Liu \bgroup et al\mbox.\egroup
  }{2016}]{liu2016not}
Liu, C.-W.; Lowe, R.; Serban, I.~V.; Noseworthy, M.; Charlin, L.; and Pineau,
  J.
\newblock 2016.
\newblock How not to evaluate your dialogue system: An empirical study of
  unsupervised evaluation metrics for dialogue response generation.
\newblock {\em EMNLP}.

\bibitem[\protect\citeauthoryear{Mi, Wang, and
  Ittycheriah}{2016}]{mi2016vocabulary}
Mi, H.; Wang, Z.; and Ittycheriah, A.
\newblock 2016.
\newblock Vocabulary manipulation for neural machine translation.
\newblock {\em Annual Meeting of the Association for Computational
  Linguistics}.

\bibitem[\protect\citeauthoryear{Mou \bgroup et al\mbox.\egroup
  }{2016}]{mou2016sequence}
Mou, L.; Song, Y.; Yan, R.; Li, G.; Zhang, L.; and Jin, Z.
\newblock 2016.
\newblock Sequence to backward and forward sequences: A content-introducing
  approach to generative short-text conversation.
\newblock {\em arXiv preprint arXiv:1607.00970}.

\bibitem[\protect\citeauthoryear{Ritter, Cherry, and
  Dolan}{2010}]{ritter2010unsupervised}
Ritter, A.; Cherry, C.; and Dolan, B.
\newblock 2010.
\newblock Unsupervised modeling of twitter conversations.

\bibitem[\protect\citeauthoryear{Ritter, Cherry, and
  Dolan}{2011}]{ritter2011data}
Ritter, A.; Cherry, C.; and Dolan, W.~B.
\newblock 2011.
\newblock Data-driven response generation in social media.
\newblock In {\em EMNLP},  583--593.
\newblock Association for Computational Linguistics.

\bibitem[\protect\citeauthoryear{Serban \bgroup et al\mbox.\egroup
  }{2016}]{serban2015building}
Serban, I.~V.; Sordoni, A.; Bengio, Y.; Courville, A.~C.; and Pineau, J.
\newblock 2016.
\newblock Building end-to-end dialogue systems using generative hierarchical
  neural network models.
\newblock In {\em AAAI},  3776--3784.

\bibitem[\protect\citeauthoryear{Serban \bgroup et al\mbox.\egroup
  }{2017}]{DBLP:conf/aaai/SerbanSLCPCB17}
Serban, I.~V.; Sordoni, A.; Lowe, R.; Charlin, L.; Pineau, J.; Courville,
  A.~C.; and Bengio, Y.
\newblock 2017.
\newblock A hierarchical latent variable encoder-decoder model for generating
  dialogues.
\newblock In {\em AAAI, February 4-9, 2017, San Francisco, California, {USA.}},
   3295--3301.

\bibitem[\protect\citeauthoryear{Shang, Lu, and Li}{2015}]{shang2015genBased}
Shang, L.; Lu, Z.; and Li, H.
\newblock 2015.
\newblock Neural responding machine for short-text conversation.
\newblock {\em ACL 2015}  1577--1586.

\bibitem[\protect\citeauthoryear{Shao \bgroup et al\mbox.\egroup
  }{2017}]{shao2017generating}
Shao, L.; Gouws, S.; Britz, D.; Goldie, A.; Strope, B.; and Kurzweil, R.
\newblock 2017.
\newblock Generating long and diverse responses with neural conversation
  models.
\newblock {\em arXiv preprint arXiv:1701.03185}.

\bibitem[\protect\citeauthoryear{Sordoni \bgroup et al\mbox.\egroup
  }{2015}]{sordoni2015neural}
Sordoni, A.; Galley, M.; Auli, M.; Brockett, C.; Ji, Y.; Mitchell, M.; Nie,
  J.-Y.; Gao, J.; and Dolan, B.
\newblock 2015.
\newblock A neural network approach to context-sensitive generation of
  conversational responses.
\newblock {\em arXiv preprint arXiv:1506.06714}.

\bibitem[\protect\citeauthoryear{Sutskever, Vinyals, and
  Le}{2014}]{sutskever2014sequence}
Sutskever, I.; Vinyals, O.; and Le, Q.~V.
\newblock 2014.
\newblock Sequence to sequence learning with neural networks.
\newblock In {\em NIPS},  3104--3112.

\bibitem[\protect\citeauthoryear{{Theano Development
  Team}}{2016}]{2016arXiv160502688short}
{Theano Development Team}.
\newblock 2016.
\newblock {Theano: A {Python} framework for fast computation of mathematical
  expressions}.
\newblock {\em arXiv e-prints} abs/1605.02688.

\bibitem[\protect\citeauthoryear{Tian \bgroup et al\mbox.\egroup
  }{2017}]{tian2017make}
Tian, Z.; Yan, R.; Mou, L.; Song, Y.; Feng, Y.; and Zhao, D.
\newblock 2017.
\newblock How to make context more useful? an empirical study on context-aware
  neural conversational models.
\newblock In {\em ACL (Volume 2: Short Papers)}, volume~2,  231--236.

\bibitem[\protect\citeauthoryear{Vinyals and Le}{2015}]{vinyals2015neural}
Vinyals, O., and Le, Q.
\newblock 2015.
\newblock A neural conversational model.
\newblock {\em arXiv preprint arXiv:1506.05869}.

\bibitem[\protect\citeauthoryear{Wang \bgroup et al\mbox.\egroup
  }{2013}]{wang2013dataset}
Wang, H.; Lu, Z.; Li, H.; and Chen, E.
\newblock 2013.
\newblock A dataset for research on short-text conversations.
\newblock In {\em EMNLP},  935--945.

\bibitem[\protect\citeauthoryear{Weaver and Tao}{2001}]{weaver2001optimal}
Weaver, L., and Tao, N.
\newblock 2001.
\newblock The optimal reward baseline for gradient-based reinforcement
  learning.
\newblock In {\em UAI},  538--545.
\newblock Morgan Kaufmann Publishers Inc.

\bibitem[\protect\citeauthoryear{Weng \bgroup et al\mbox.\egroup
  }{2017}]{weng2017neural}
Weng, R.; Huang, S.; Zheng, Z.; Dai, X.; and Chen, J.
\newblock 2017.
\newblock Neural machine translation with word predictions.
\newblock {\em arXiv preprint arXiv:1708.01771}.

\bibitem[\protect\citeauthoryear{Wu \bgroup et al\mbox.\egroup
  }{2017}]{DBLP:conf/acl/WuWXZL17}
Wu, Y.; Wu, W.; Xing, C.; Zhou, M.; and Li, Z.
\newblock 2017.
\newblock Sequential matching network: {A} new architecture for multi-turn
  response selection in retrieval-based chatbots.
\newblock In {\em ACL 2017, Vancouver, Canada, July 30 - August 4, Volume 1:
  Long Papers},  496--505.

\bibitem[\protect\citeauthoryear{Xing \bgroup et al\mbox.\egroup
  }{2016}]{xing2016topic}
Xing, C.; Wu, W.; Wu, Y.; Liu, J.; Huang, Y.; Zhou, M.; and Ma, W.-Y.
\newblock 2016.
\newblock Topic aware neural response generation.
\newblock {\em AAAI}.

\bibitem[\protect\citeauthoryear{Xu \bgroup et al\mbox.\egroup
  }{2017}]{xu-EtAl:2017:EMNLP2017}
Xu, Z.; Liu, B.; Wang, B.; SUN, C.; Wang, X.; Wang, Z.; and Qi, C.
\newblock 2017.
\newblock Neural response generation via gan with an approximate embedding
  layer.
\newblock In {\em EMNLP 2017}.

\bibitem[\protect\citeauthoryear{Yan, Song, and Wu}{2016}]{yan2016learning}
Yan, R.; Song, Y.; and Wu, H.
\newblock 2016.
\newblock Learning to respond with deep neural networks for retrieval-based
  human-computer conversation system.
\newblock In {\em SIGIR},  55--64.
\newblock ACM.

\bibitem[\protect\citeauthoryear{Zeiler}{2012}]{zeiler2012adadelta}
Zeiler, M.~D.
\newblock 2012.
\newblock Adadelta: an adaptive learning rate method.
\newblock {\em arXiv preprint arXiv:1212.5701}.

\bibitem[\protect\citeauthoryear{Zhao, Zhao, and
  Esk{\'{e}}nazi}{}]{DBLP:conf/acl/ZhaoZE17}
Zhao, T.; Zhao, R.; and Esk{\'{e}}nazi, M.
\newblock Learning discourse-level diversity for neural dialog models using
  conditional variational autoencoders.
\newblock In {\em ACL 2017 Long Papers}.

\bibitem[\protect\citeauthoryear{Zhou \bgroup et al\mbox.\egroup
  }{2017}]{zhou2017emotional}
Zhou, H.; Huang, M.; Zhang, T.; Zhu, X.; and Liu, B.
\newblock 2017.
\newblock Emotional chatting machine: Emotional conversation generation with
  internal and external memory.
\newblock {\em arXiv preprint arXiv:1704.01074}.

\end{thebibliography}
	\end{CJK*}
\end{document}